\documentclass[letterpaper]{article} % DO NOT CHANGE THIS
\usepackage[draft]{aaai2027} % DO NOT CHANGE THIS
\usepackage[hyphens]{url} % DO NOT CHANGE THIS
\usepackage{graphicx} % DO NOT CHANGE THIS
\usepackage{natbib} % DO NOT CHANGE THIS AND DO NOT ADD OPTIONS
\usepackage{caption} % DO NOT CHANGE THIS AND DO NOT ADD OPTIONS
\usepackage{makecell}
\usepackage{amsmath}

\usepackage{booktabs}
\usepackage{multirow}

\usepackage{array}
\usepackage{tabularx}

\newcolumntype{Y}{>{\centering\arraybackslash}X}
\newcommand{\metrichead}[1]{%
  \parbox[c]{\linewidth}{\centering\footnotesize #1}%
}
\title{QUMem: Personalized Memory for Query-Conditioned User-State Inference in LLM Agents}
\author{
Heng Wang$^{1,2,*}$,
Yifei Li$^{1,2,*}$,
Lingling Zhang$^{1,2,\dagger}$,
Pengyu Li$^{1,2}$,
Xinyu Che$^{1,2}$,
Xinyu Zhang$^{1,2}$,
Zesheng Yang$^{1,2}$
}

\affiliations{
$^{1}$School of Computer Science and Technology, Xi'an Jiaotong University\\
$^{2}$MOE KLNN Lab, Xi'an Jiaotong University\\
$^{*}$Equal contribution. \quad $^{\dagger}$Corresponding author.\\
\texttt{wangheng8280@stu.xjtu.edu.cn}
}

\begin{document}

\maketitle

\begin{abstract}
Large language model (LLM) agents increasingly use external memory systems to support personalization by drawing on long and evolving interaction histories, in which user preferences may be distributed across time, change with context, and conflict with earlier evidence. However, existing systems face three limitations: fixed-turn, fixed-token, or session-based boundaries can mix unrelated dialogue or split an event from its causes, decisions, and outcomes; storing multiple pieces of user information from the same interaction as a single memory binds together items that serve different functions and should be independently retrievable; and treating the current task as a single top-$k$ retrieval query can return fragments that are individually relevant but fail to jointly capture preference evolution, temporal validity, and contextual applicability. We introduce \textsc{QUMem}, a structured memory framework for query-conditioned user-state inference. \textsc{QUMem} first segments interaction histories into variable-length episodes according to semantic continuity, then decomposes each episode into independently retrievable factual, preference, and transferable insight memories while preserving temporal positions and source evidence. At inference time, three sequential agents identify task-specific information needs, plan multi-query retrieval over the typed memory stores, and jointly infer a temporally and contextually valid user state for downstream response generation. \textsc{QUMem} achieves state-of-the-art performance on both PersonaMem and KnowU-Bench, demonstrating the effectiveness of query-conditioned user-state inference for long-term personalization.
\end{abstract}

% Optional anonymous resource links belong between the abstract and main text.
% \begin{links}
%     \link{Code}{https://aaai.org/example/code}
%     \link{Datasets}{https://aaai.org/example/datasets}
%     \link{Extended version}{https://aaai.org/example/extended-version}
% \end{links}

\section{Introduction}
Recent advances in reasoning, planning, tool use, and environment interaction have transformed large language models (LLMs) into agents capable of completing multi-step tasks with increasing autonomy \cite{yao2023reactsynergizingreasoningacting,wang2023voyager,qin2025uitarspioneeringautomatedgui}. As these agents increasingly serve as persistent personal assistants across sessions, they must interact with the same user over time, carry forward unfinished tasks and constraints, and adapt their behavior based on prior feedback. Personalized memory supports this continuity by retaining evidence from past interactions, but effective personalization requires more than recalling isolated facts. An assistant must integrate preferences distributed across interactions, distinguish currently valid tendencies from context-dependent or obsolete ones, interpret preference changes, and transfer prior decision rationales to new situations. Personalized memory systems are therefore a core requirement for persistent assistance.

\begin{figure}[t]
    \centering
    \includegraphics[
        width=\columnwidth
    ]{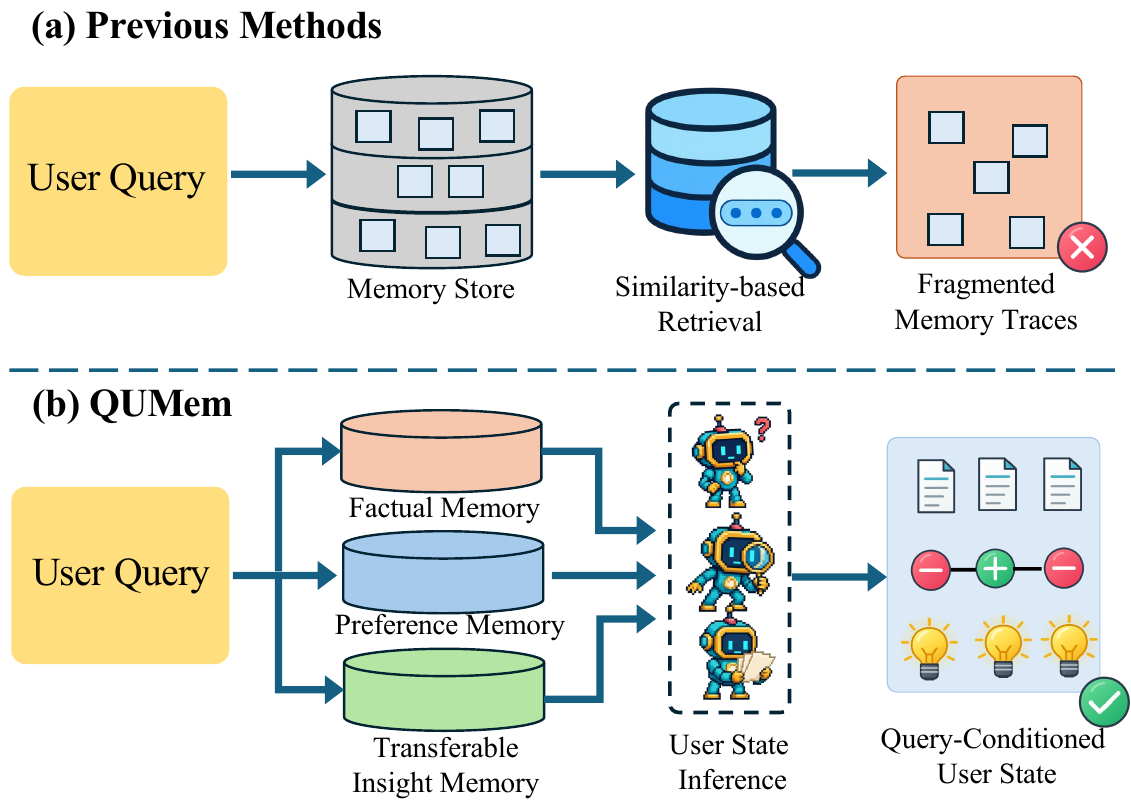}
    \caption{Comparison between previous method and
\textsc{QUMem}.  Previous methods retrieve memory fragments based on their similarity to the current query, whereas \textsc{QUMem} retrieves task-relevant evidence from typed memory stores and jointly interprets it to infer the user's current, contextually valid state.}
    \label{fig:framework-overview}
\end{figure}

The goal of a memory system is to use historical information to support personalized judgments and actions for the current task. For persistent agents, the system must identify information from an evolving history that is both relevant to the current task and applicable in the current context. Existing personalized memory frameworks pursue this goal by improving memory storage and retrieval \cite{xu2025amemagenticmemoryllm,chhikara2025mem0}. Despite this progress, existing methods still face three limitations. First, memory units defined by fixed turn counts, token counts, or session boundaries may absorb unrelated dialogue or split an event from its causes, decisions, and outcomes, making it difficult for subsequent retrieval to repair the disrupted event-level context \cite{zhang2025survey}. Second, a single user interaction often contains multiple pieces of information with different semantics and functions that could be reused independently. Storing them together as a single memory binds them during subsequent retrieval, making it difficult for the system to retrieve only the part relevant to the current query. For example, a user may state both ``I prefer concise implementations'' and ``the current project must remain compatible with Python 3.9'' in the same programming conversation. The former may inform future programming tasks, whereas the latter applies only to the current project; they should not be stored as an indivisible memory that must be retrieved as a whole. Third, the current task is not always an effective similarity query, particularly when user preferences evolve. Independent top-$k$ retrieval may return fragments that are individually relevant but do not collectively cover the user's current state, the reasons for state transitions, or the temporal validity of the evidence.

These limitations indicate that a memory system should preserve event-level context during construction, distinguish the functional roles of different evidence during representation, and jointly assess relevance, temporal validity, and contextual applicability at query time. To this end, the system must perform user-state inference by retrieving and jointly interpreting evidence distributed throughout the interaction history in light of the current query. Here, the ``user state'' is a structured representation of user information that is supported by historical evidence, relevant to the current query, and applicable in the current context. The central problem in long-term personalization therefore lies not only in finding relevant memories, but also in using them to infer a task-relevant user state that is temporally and contextually valid. Figure~\ref{fig:framework-overview} illustrates how this process differs from conventional memory systems.

Motivated by these observations, we propose \textsc{QUMem}, a structured personalized memory framework for query-conditioned user-state inference. To address the loss of event context caused by fixed boundaries, \textsc{QUMem} organizes long interaction histories into granularity-adaptive dialogue episodes based on semantic continuity, allowing the causes, decisions, feedback, and outcomes of the same event to be interpreted together. To avoid coupling multiple pieces of user information within a single interaction, the framework further decomposes each episode into independently retrievable factual memories, preference memories that encode preferences and constraints, and transferable insight memories. Downstream tasks can then select and combine historical evidence with different functions as needed. Each memory retains its temporal position and source links, supporting assessment of its temporal validity and tracing of its provenance.

Given a current query, \textsc{QUMem} uses three sequential agents for information-need identification, retrieval planning, and user-state inference to progressively transform the task objective into a query-relevant representation of user information. The three stages determine, respectively, what the current task requires the system to verify, from which typed memories the evidence should be retrieved, and what current state the evidence jointly supports. Unlike a single top-$k$ retrieval operation that directly uses the original query, this process explicitly separates evidence requirements, evidence acquisition, and evidence interpretation, enabling evidence distributed across time to be jointly interpreted for the current task.
Our contributions are as follows:
\begin{itemize}
    \item We propose \textsc{QUMem}, a structured memory framework for long-term personalization. It first organizes event-level dialogue episodes based on semantic continuity and then decomposes the user information in each episode into complementary and independently retrievable factual, preference, and transferable insight memories, preserving event context while supporting fine-grained information reuse.
    \item We introduce a query-conditioned three-agent user-state inference mechanism that decomposes information-need identification, retrieval planning over typed memories, and evidence-based state inference into sequential stages, enabling the joint interpretation of distributed, temporally separated, and context-dependent user evidence.
    \item We evaluate \textsc{QUMem} on PersonaMem and KnowU-Bench. It achieves strong performance on both benchmarks, supporting the effectiveness of the method.
    % [NEEDS EVIDENCE: datasets and task settings], showing [NEEDS EVIDENCE: headline results against named baselines and the main ablation findings].
\end{itemize}

\section{Related Work}
\begin{figure*}[t]
    \centering
    \includegraphics[width=\textwidth]{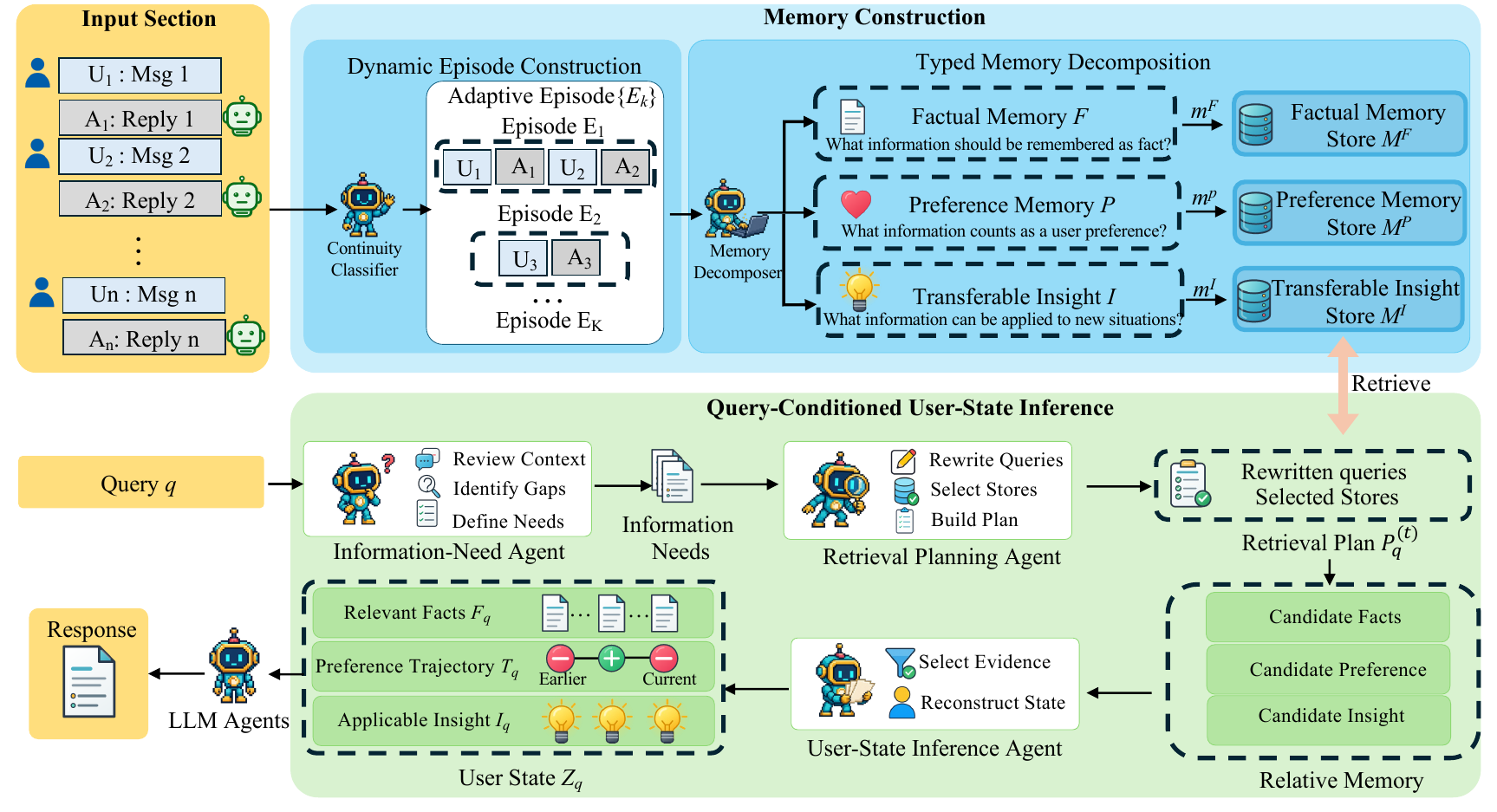}
    \caption{Overview of \textsc{QUMem}. The framework constructs semantically coherent dialogue episodes, decomposes them into factual, preference, and transferable insight memories, and uses three agents to infer a query-conditioned user state.}
    \label{fig:method-framework}
\end{figure*}
\subsection{Personalized Memory for LLM Agents}
For long-term personalization, A-MEM organizes experiences as linked, structured notes, Mem0 maintains salient user information through explicit addition, update, deletion, and retention operations, and Zep uses a temporal knowledge graph to track fact validity and conflicts \cite{xu2025amemagenticmemoryllm,chhikara2025mem0,rasmussen2025zeptemporalknowledgegraph}. Work on memory granularity and structure includes SeCom, which combines topic-aware segmentation with compressed retrieval; Reflective Memory Management (RMM), which builds and retrieves multi-granularity summaries through reflection; HyperMem, which organizes topics, events, and facts in a hierarchical hypergraph; and HingeMem, which couples boundary detection with query-adaptive retrieval \cite{pan2025memory,tan-etal-2025-prospect,yue-etal-2026-hypermem,10.1145/3774904.3792089}. Recent studies further examine query-conditioned user modeling and memory validity. Park et al. retrieve user history to construct task-specific profiles, while Memory Retrieval for Changing Preferences selects historical evidence relevant to evolving preferences. RaMem, STALE, and MemORAI respectively address contextual reinstatement, stale-memory detection, and provenance-aware adaptive retrieval \cite{park-etal-2026-learning,qin2026memoryretrievalchangingpreferences,yang2026ramemcontextualreinstatementlongterm,chao2026stalellmagentsknow,van-etal-2026-memorai}. \textsc{QUMem} focuses on connecting semantically coherent episodes to factual, preference, and transferable insight memories, together with their temporal positions and source evidence. This structure supports the inference of a traceable, query-conditioned user state that is applicable to the current task context.

\subsection{Retrieval-Augmented Generation}
Retrieval-augmented generation (RAG) augments generation with evidence retrieved from external non-parametric knowledge sources \cite{lewis2020retrieval}. Subsequent work uses query rewriting, pre-retrieval planning, and question decomposition to transform complex requests into retrieval-oriented queries or subquestions \cite{ma-etal-2023-query,lee-etal-2024-planrag,petcu-etal-2026-query}. SAFARI, Omni-RAG, and DeepSieve further select among knowledge sources and generate source-specific queries or recursively route subquestions \cite{wang-etal-2023-large,chen-etal-2025-towards-omni,guo-etal-2026-deepsieve}; Chain-of-Note and MASS-RAG filter, assess, and synthesize distributed evidence through explicit reading notes or role-specialized agents \cite{yu-etal-2024-chain,xiao-etal-2026-mass}. These approaches primarily address access to external knowledge, whereas \textsc{QUMem} operates over a continually evolving user interaction history. It first identifies the user information that must be verified for the current task, routes rewritten queries to typed memory stores, and organizes the retrieved evidence by provenance, temporal position, and contextual applicability into a query-conditioned user state for downstream personalized decisions.
\section{Method}

We propose QUMem, a memory system for long-term personalization that organizes interaction histories as evidence for query-time user-state inference. QUMem comprises three core modules: Dynamic Episode Construction, Typed Memory Decomposition, and Query-Conditioned User-State Inference.Figure~\ref{fig:method-framework} presents the overall framework.

\subsection{Problem Formulation}

Let $\mathcal{H}=(h_1,\ldots,h_T)$ denote a user's chronologically
ordered multi-session interaction history, and let $q$ denote the
current query. We formulate the task as
\[
    \mathcal{Z}_q = \Phi_{\mathrm{mem}}(\mathcal{H},q),
    \qquad
    \widehat{y}_q = \Psi(q,\mathcal{Z}_q),
\]
where $\mathcal{Z}_q$ is the history-grounded user state relevant to
$q$, and $\widehat{y}_q$ is the resulting personalized output.
QUMem instantiates $\Phi_{\mathrm{mem}}$, while $\Psi$ denotes the
downstream response model.

\subsection{Dynamic Episode Construction}

Events in interaction histories span varying numbers of turns. Consequently, dialogue episodes constructed using fixed turn counts, token counts, or session boundaries may not align with the underlying event structure. Episodes that are too coarse may mix entities, preferences, and feedback from unrelated events, whereas overly fine-grained episodes may separate the context, decisions, and outcomes of the same event. Because these boundary errors occur before memories are written, subsequent retrieval cannot readily recover the disrupted event context. We therefore maintain a candidate episode based on the semantic continuity between adjacent user utterances and finalize it when that continuity is broken.

The system maintains an open candidate episode $\widetilde{E}_k$ throughout the interaction. The first user utterance $x_1$, together with its ensuing assistant response, initializes $\widetilde{E}_1$. For each subsequent user utterance $x_t$, the continuity classifier $f_{\theta}$ determines whether it continues the same event, task, or decision process as the preceding user utterance $x_{t-1}$:

\[
  c_t=f_{\theta}(x_{t-1},x_t),
  \qquad
  c_t\in\{0,1\},
  \qquad t=2,\ldots,N.
\]
Here, $c_t=1$ indicates that $x_t$ and $x_{t-1}$ belong to the same event, whereas $c_t=0$ indicates that $x_t$ starts a new event, placing a semantic boundary between them. The classifier receives only the two adjacent user utterances. The intervening assistant response is excluded from the boundary decision but remains in the current candidate episode as interaction context.

For a history containing $N$ user utterances, the system makes $N-1$ boundary decisions. Because each decision is a well-defined binary classification task that does not require open-ended generation, we implement $f_{\theta}$ as a lightweight model fine-tuned for continuity classification. This reduces the computational cost and inference latency that would arise from repeatedly invoking a large generative model.

When $c_t=1$, the system appends $x_t$ and its corresponding assistant response to the current candidate episode $\widetilde{E}_k$. When $c_t=0$, it finalizes $\widetilde{E}_k$ as a dialogue episode $E_k$ and passes it to the subsequent typed memory decomposition module, while initializing a new candidate episode $\widetilde{E}_{k+1}$ with $x_t$. When processing a finite interaction history, the system also finalizes and submits any remaining open candidate episode after the final interaction.
\subsection{Typed Memory Decomposition}
The preceding Dynamic Episode Construction stage determines which interactions should be interpreted jointly, thereby preserving complete event context whenever possible. However, an episode may still contain multiple pieces of user information that differ in semantics and function and can be reused independently. Storing the entire episode as a single memory couples these pieces during subsequent retrieval, making it difficult to retrieve only those relevant to the current query. We therefore retain the episode as context for joint interpretation and evidence verification while decomposing the user information it contains into independently retrievable atomic memories with explicit types.

This type system corresponds to three basic roles of historical evidence in long-term personalization: recalling user experiences, conditioning subsequent decisions on specific preferences and constraints, and transferring decision principles reflected in prior interactions to new contexts\cite{jiang2025knowmerespondme}. 

\textbf{Factual memories ($F$)} record concrete user experiences, behaviors, activities, states, or events without inferring broader tendencies. 

\textbf{Preference memories ($P$)} record a user's choices, tendencies, requirements, and constraints concerning specific objects or contexts, together with directly associated rationales. A preference memory is not assumed to remain valid indefinitely; its temporal position may reflect either a persistent tendency or a particular stage in preference evolution. 

\textbf{Transferable insight memories ($I$)} abstract user-specific decision principles from concrete choices, feedback, and their rationales. The resulting principles can be applied to new objects or contexts but must remain grounded in concrete interaction evidence. For example, ``the user joined a swimming club,'' ``the user prefers swimming classes with a supportive environment,'' and ``the user tends to choose low-pressure exercise environments'' constitute a factual memory, a preference memory, and a transferable insight memory, respectively.

For each episode, we invoke the same LLM-based memory decomposer $g_{\phi}$ three times, conditioning each invocation on one memory type $d$:

\[
  g_{\phi}(E_k,d)
  =
  \{m_{k,j}^{d}\}_{j=1}^{n_k^{d}},
  \qquad
  d\in\mathcal{D}=\{F,P,I\}.
\]
Each atomic memory expresses a single independently retrievable item of user information. An episode may yield multiple memories of the same or different types, and the same source interactions may support memories at different levels of abstraction.

Each memory is represented as

\[
  m_{k,j}^{d}
  =
  \left(
  v_{k,j}^{d},
  d,
  p_{k,j}^{d},
  \mathcal{E}_{k,j}^{d}
  \right),
  \qquad
  p_{k,j}^{d}
  =
  \max_{e\in\mathcal{E}_{k,j}^{d}}\operatorname{pos}(e).
\]
Here, $v_{k,j}^{d}$ is the memory content, $\mathcal{E}_{k,j}^{d}$ is the set of interaction turns supporting the memory, and $p_{k,j}^{d}$ is the temporal position of the latest such turn. The index $k$ also links the memory to its source episode $E_k$. The extracted memories are stored by type in $\mathcal{M}^{d}=\bigcup_k g_{\phi}(E_k,d)$. Thus, source episodes preserve complete event context, atomic memories support independent retrieval, and typed memory stores provide an interface for dynamic routing at query time.
\subsection{Query-Conditioned User-State Inference}
\begin{table*}[t]
\centering
\small
\setlength{\tabcolsep}{1.8pt}
\begin{tabularx}{\textwidth}{@{}lll*{8}{Y}@{}}
\toprule
\makecell[c]{Base\\Model}
& \makecell[c]{Context\\Length}
& \makecell[c]{Method}
& \metrichead{Recall\\user-shared\\facts}
& \metrichead{Suggest\\new\\ideas}
& \metrichead{Acknowledge\\latest user\\preferences}
& \metrichead{Track full\\preference\\evolution}
& \metrichead{Revisit\\reasons behind\\preference updates}
& \metrichead{Provide\\preference-aligned\\recommendation}
& \metrichead{Generalize to\\new scenarios}
& \metrichead{Overall} \\
\midrule
\multirow{16}{*}{%
  \rotatebox[origin=c]{90}{GPT-4o-mini}%
}
& \multirow{4}{*}{32K} & A-MEM & 47.95 & 30.11 & -- & 44.60 & 71.72 & 52.73 & 17.54 & 45.84 \\
& & Mem0 & \underline{67.81} & \underline{36.56} & -- & 47.48 & \underline{83.84} & 65.45 & \underline{63.16} & \underline{60.10} \\
& & Zep & 45.89 & 31.18 & -- & \underline{51.08} & 70.71 & \underline{69.09} & 31.58 & 49.75 \\
& & \textsc{QUMem} & \textbf{69.18} & \textbf{49.46} & -- & \textbf{54.68} & \textbf{84.85} & \textbf{76.36} & \textbf{73.68} & \textbf{66.38} \\
\cmidrule(lr){2-11}
& \multirow{4}{*}{128K} & A-MEM & 59.65 & \underline{31.27} & 50.12 & 46.04 & 65.06 & 44.99 & 27.23 & 45.65 \\
& & Mem0 & \underline{75.44} & 30.31 & \underline{62.47} & \underline{54.84} & \underline{79.55} & \underline{63.04} & \underline{44.60} & \underline{56.58} \\
& & Zep & 59.06 & 30.12 & 45.15 & 51.61 & 65.80 & 51.29 & 31.92 & 45.76 \\
& & \textsc{QUMem} & \textbf{84.21} & \textbf{36.29} & \textbf{66.28} & \textbf{64.52} & \textbf{89.96} & \textbf{72.21} & \textbf{66.67} & \textbf{64.61} \\
\cmidrule(lr){2-11}
& \multirow{4}{*}{1M} & A-MEM & 51.39 & 27.24 & 45.05 & 42.67 & 56.17 & 34.64 & 26.78 & 38.22 \\
& & Mem0 & \underline{75.00} & \underline{27.65} & \underline{58.20} & 46.22 & \underline{77.87} & \underline{44.64} & \underline{36.95} & \underline{47.76} \\
& & Zep & 52.08 & 27.37 & 42.71 & \underline{49.33} & 63.40 & 40.36 & 26.44 & 39.38 \\
& & \textsc{QUMem} & \textbf{87.50} & \textbf{33.84} & \textbf{61.20} & \textbf{60.00} & \textbf{89.36} & \textbf{58.57} & \textbf{51.19} & \textbf{56.17} \\
\cmidrule(lr){2-11}
& \multirow{4}{*}{All} & A-MEM & 53.36 & 29.00 & 47.74 & 44.68 & 62.69 & 41.37 & 26.02 & 42.35 \\
& & Mem0 & \underline{72.89} & \underline{29.30} & \underline{60.47} & 50.64 & \underline{79.60} & \underline{55.70} & \underline{42.48} & \underline{52.99} \\
& & Zep & 52.71 & 28.70 & 44.00 & \underline{50.78} & 65.67 & 48.25 & 29.03 & 43.31 \\
& & \textsc{QUMem} & \textbf{80.48} & \textbf{35.87} & \textbf{63.89} & \textbf{61.13} & \textbf{88.89} & \textbf{66.96} & \textbf{59.29} & \textbf{61.02} \\
\midrule
\multirow{16}{*}{%
  \rotatebox[origin=c]{90}{Gemini-3.5-flash}%
}
& \multirow{4}{*}{32K} & A-MEM & 52.74 & 39.78 & -- & 53.96 & 78.79 & 63.64 & 26.32 & 53.82 \\
& & Mem0 & \underline{71.92} & \underline{47.31} & -- & 56.83 & \underline{88.89} & 74.55 & \underline{75.44} & \underline{67.91} \\
& & Zep & 50.68 & 40.86 & -- & \underline{59.71} & 78.79 & \underline{78.18} & 43.86 & 57.89 \\
& & \textsc{QUMem} & \textbf{73.29} & \textbf{60.22} & -- & \textbf{63.31} & \textbf{89.90} & \textbf{83.64} & \textbf{82.46} & \textbf{73.51} \\
\cmidrule(lr){2-11}
& \multirow{4}{*}{128K} & A-MEM & 64.33 & \underline{41.12} & 65.59 & 55.13 & 73.61 & 55.87 & 39.44 & 57.06 \\
& & Mem0 & \underline{78.95} & 40.15 & \underline{75.98} & \underline{63.64} & \underline{85.50} & \underline{72.49} & \underline{58.22} & \underline{66.92} \\
& & Zep & 63.74 & 39.96 & 60.97 & 60.41 & 74.35 & 62.18 & 45.07 & 57.32 \\
& & \textsc{QUMem} & \textbf{86.55} & \textbf{46.72} & \textbf{78.87} & \textbf{72.43} & \textbf{92.94} & \textbf{80.23} & \textbf{77.46} & \textbf{73.89} \\
\cmidrule(lr){2-11}
& \multirow{4}{*}{1M} & A-MEM & 56.25 & 36.59 & 60.94 & 51.56 & 65.96 & 45.36 & 38.98 & 49.66 \\
& & Mem0 & \underline{78.47} & \underline{37.00} & \underline{72.53} & 55.11 & \underline{84.26} & \underline{55.71} & \underline{50.51} & \underline{58.56} \\
& & Zep & 56.94 & 36.73 & 58.59 & \underline{58.22} & 72.34 & 51.43 & 38.64 & 50.79 \\
& & \textsc{QUMem} & \textbf{89.58} & \textbf{44.02} & \textbf{75.00} & \textbf{68.44} & \textbf{92.77} & \textbf{68.57} & \textbf{64.75} & \textbf{66.57} \\
\cmidrule(lr){2-11}
& \multirow{4}{*}{All} & A-MEM & 58.13 & 38.57 & 63.40 & 53.76 & 71.48 & 52.19 & 37.88 & 53.44 \\
& & Mem0 & \underline{76.57} & \underline{38.94} & \underline{74.36} & \underline{59.57} & \underline{85.57} & \underline{65.79} & \underline{55.93} & \underline{63.29} \\
& & Zep & 57.48 & 38.27 & 59.85 & \underline{59.57} & 74.30 & 59.06 & 41.59 & 54.46 \\
& & \textsc{QUMem} & \textbf{83.30} & \textbf{46.19} & \textbf{77.05} & \textbf{69.36} & \textbf{92.37} & \textbf{75.73} & \textbf{71.33} & \textbf{70.58} \\
\bottomrule
\end{tabularx}
\caption{Performance on the PersonaMem benchmark. The best performance is highlighted in \textbf{bold}, and the second-best is \underline{underlined}.}
\label{tab:personamem-results}
\end{table*}
The current query typically specifies the task objective but does not explicitly state which user information should be verified from history. Directly using the query for retrieval collapses three distinct decisions into a single similarity search: what historical evidence the current task requires, which memory stores should provide that evidence, and what user state the evidence jointly supports. \textsc{QUMem} therefore uses three collaborating agents with complementary responsibilities: the Information-Need Agent, the Retrieval Planning Agent, and the User-State Inference Agent. Together, they address these decisions and infer a task-relevant user state from the long-term interaction history.

\paragraph{Information-Need Agent $A_1$}
Given the current query $q$ and the available task context, $A_1$ identifies the information that must be verified from history and explains why it matters to user-state inference. If the task concerns preferences that may have changed over time, the information needs cover earlier and later expressions of those preferences, the contexts in which each applies, and any reasons for change supported by historical evidence. At this stage, $A_1$ does not prescribe specific retrieval queries or memory stores, thereby separating information-need identification from retrieval planning.

\paragraph{Retrieval Planning Agent $A_2$}
$A_2$ rewrites each information need as one or more self-contained queries and selects one or more of the factual, preference, and transferable insight memory stores for each query. The retrieval plan for the current query is
\[
\mathcal{P}_q
=
\left\{
\left(\widetilde q_j,\mathcal{D}_j\right)
\right\}_{j=1}^{J},
\qquad
\emptyset\neq\mathcal{D}_j\subseteq\mathcal{D}.
\]
Here, $\widetilde q_j$ is the $j$-th rewritten query, $\mathcal{D}_j$ is the set of memory types selected for that query, and $J$ is the number of rewritten queries. According to $\mathcal{P}_q$, the system queries the selected typed memory stores, then merges and deduplicates the candidate memories returned across queries and stores. The resulting candidates serve as evidence for user-state inference.

\paragraph{User-State Inference Agent $A_3$}
Given the current query $q$ and the retrieved candidate memories, $A_3$ selects task-relevant information and organizes it into the query-conditioned user state
\[
\mathcal{Z}_q
=
\left(
\mathcal{F}_q,
\mathcal{T}_q,
\mathcal{I}_q
\right).
\]
Here, $\mathcal{F}_q$ organizes facts relevant to the current task chronologically to describe how the corresponding experiences and states have developed over time; $\mathcal{T}_q$ characterizes the evolution of the user's preferences and identifies those applicable in the current context; and $\mathcal{I}_q$ specifies how decision principles abstracted from history apply to the current task.

The inferred user state $\mathcal{Z}_q$ is provided to the downstream response model together with the current query $q$ to generate a personalized response:
\[
\widehat y_q=\Psi(q,\mathcal{Z}_q).
\]
Here, $\Psi$ denotes the downstream response model. By providing a structured representation of the task-relevant user state, $\mathcal{Z}_q$ enables the model to interpret the current query in light of relevant user information from the interaction history.

\section{Experiments}

\subsection{Evaluation Settings}
\paragraph{Datesets.}

To evaluate whether our method can reconstruct task-relevant and temporally valid user states from long-term interaction histories,We conduct experiments on PersonaMem \cite{jiang2025knowmerespondme}. PersonaMem evaluates dynamic user modeling in long-horizon dialogues, requiring systems to recall user information, track evolving preferences, and apply the inferred user state to personalized responses in both familiar and new contexts. To further evaluate the generalizability of our method, we conduct additional experiments on KnowU-Bench \cite{chen2026knowubench}. KnowU-Bench evaluates whether personalized mobile agents can infer user preferences and constraints from behavioral histories and translate them into concrete actions.

\paragraph{Baselines.}
We selected three representative long-term memory systems as the primary baselines. A-MEM \cite{xu2025amemagenticmemoryllm} draws on the Zettelkasten method and organizes interactions into structured memories that can be dynamically linked and updated. Mem0 \cite{chhikara2025mem0} extracts user-related information from conversations and maintains persistent memory by adding, updating, and deleting entries. Zep \cite{rasmussen2025zeptemporalknowledgegraph} uses a time-aware knowledge graph to manage facts and their validity and constructs the memory context by combining semantic, textual, and graph-based retrieval.

\paragraph{Evaluation Metrics.}
PersonaMem formulates evaluation as a four-choice response-selection task, for which we use accuracy as the evaluation metric. For KnowU-Bench, we use Success Rate (SR), Average Score, and Interaction Efficiency (IE) following the official evaluation protocol.

\paragraph{Experimental Details.}
The Dynamic Episode Construction module used a semantic continuity classifier fine-tuned from the Qwen3.5-4B model. We constructed the corresponding supervised fine-tuning dataset from LoCoMo dialogues \cite{maharana-etal-2024-evaluating}. The PersonaMem experiments used GPT-4o-mini and Gemini-3.5-flash as the base model. The KnowU-Bench experiments used GPT-4o-mini as the base model.  We set the retrieval depth to $k=5$. Additional implementation details are provided in the appendix.

\subsection{Main Results}
\paragraph{PersonaMem.}
Table~\ref{tab:personamem-results} shows that \textsc{QUMem} outperforms all evaluated baselines across every comparable context configuration and query category under both base models. With GPT-4o-mini, \textsc{QUMem} improves overall accuracy from 52.99\% to 61.02\%, while under Gemini-3.5-flash it improves the strongest baseline from 63.29\% to 70.58\%. The consistent ranking across the two base models suggests that the effectiveness of the memory framework is not tied to a particular downstream model.

The performance pattern across query categories further clarifies where these gains arise. The largest advantages occur in tracking complete preference evolution, producing preference-aligned recommendations, and generalizing prior user information to new scenarios. These tasks require evidence from different times or contexts to be interpreted jointly. By contrast, the improvement on recognizing the latest preference is smaller, suggesting that the main benefit of \textsc{QUMem} goes beyond retrieving a recent preference statement. This pattern is consistent with the intended role of query-conditioned user-state inference in organizing temporally and contextually distributed evidence.

The overall margin over the strongest baseline widens as the interaction history grows, indicating that the structured memory and inference pipeline becomes increasingly useful when relevant evidence is dispersed across longer contexts. Nevertheless, absolute performance still declines in the 1M-token setting. Suggesting new ideas also remains the weakest category despite the consistent improvement over the baselines. These results highlight a remaining distinction between accurately inferring user preferences and generating responses that are both novel and preference-aligned.
\paragraph{KnowU-Bench.}
\begin{table}[t]
\centering
\small
\setlength{\tabcolsep}{4pt}
\begin{tabular}{lccccccc}
\toprule
& \multicolumn{2}{c}{Easy} & \multicolumn{2}{c}{Hard} & \multicolumn{3}{c}{Overall} \\
\cmidrule(lr){2-3}\cmidrule(lr){4-5}\cmidrule(lr){6-8}
Method & SR &  Score & SR &  Score & SR & Score & IE \\
\midrule
A-MEM & 14.0 & 0.41 & 4.7 & 0.34 & 9.3 & 0.38 & 0.34 \\
Mem0 & 18.6 & 0.48 & 7.0 & 0.40 & 12.8 & 0.44 & 0.40 \\
Zep & 16.3 & 0.44 & 7.0 & 0.37 & 11.6 & 0.41 & 0.37 \\
\textsc{QUMem} & \textbf{23.3} & \textbf{0.56} & \textbf{11.6} & \textbf{0.49} & \textbf{17.4} & \textbf{0.53} & \textbf{0.49} \\
\bottomrule
\end{tabular}
\caption{Performance on the KnowU-Bench benchmark.}
\label{tab:knowu-personalized}
\end{table}

Table~\ref{tab:knowu-personalized} shows that \textsc{QUMem} achieves the best performance among the evaluated methods across all overall metrics and both difficulty subsets, improving overall success rate by 4.6 percentage points over the strongest baseline. Unlike PersonaMem's response-selection setting, KnowU-Bench requires an agent to translate historical preferences and constraints into concrete actions. The consistent improvement therefore provides evidence that the benefits of query-conditioned user-state inference extend to personalized task execution. However, the relatively low success rate, particularly on hard tasks, indicates that reliable end-to-end personalized execution remains challenging.
\subsection{Ablation Studies}
Unless otherwise specified, all experiments are conducted on the PersonaMem benchmark using GPT-4o-mini as the base model.

\paragraph{Component Ablations}
\begin{table}[t]
\centering
\footnotesize
\setlength{\tabcolsep}{3pt}
\begin{tabular}{@{}lcccc@{}}
\toprule
\multirow{2}{*}{Ablation Setting} & \multicolumn{3}{c}{Context Length} & \multirow{2}{*}{Overall} \\
\cmidrule(lr){2-4}
& 32K & 128K & 1M & \\
\midrule
\textsc{QUMem} & \textbf{66.38} & \textbf{64.61} & \textbf{56.17} & \textbf{61.02} \\
\makecell[l]{w/o Episode Construction} & 64.69 & 62.05 & 53.25 & 58.38 \\
\makecell[l]{w/o Memory Decomposition} & 63.67 & 60.69 & 52.02 & 57.11 \\
\makecell[l]{w/o  Reconstruction} & 61.97 & 58.12 & 49.18 & 54.51 \\
\bottomrule
\end{tabular}
\caption{Ablation on key components.}
\label{tab:component-ablation}
\end{table}
We conducted component ablations under all three PersonaMem context configurations. Implementation details are provided in the appendix.
Table~\ref{tab:component-ablation} shows that removing any of the three
components consistently degrades performance, confirming that they make
complementary contributions to the full framework. The query-time
User-State Reconstruction pipeline has the strongest overall effect,
highlighting the importance of task-driven evidence acquisition and joint
interpretation. Typed Memory Decomposition and Dynamic Episode Construction
also provide consistent benefits by distinguishing the roles of historical
evidence and preserving coherent event context, respectively. Moreover, the
impact of all three components becomes more pronounced as interaction
histories grow, supporting their intended role in handling increasingly
dispersed evidence and mixed contexts.

\paragraph{Sensitivity to Retrieval Depth.}
We examined the effect of retrieval depth $k$ on final performance, where $k$ denotes the number of top-ranked candidate memories returned by each retrieval. Using GPT-4o-mini, we kept all other settings fixed and varied $k\in\{3,5,10\}$. Results for each category were aggregated over all available instances across the three context configurations; because the 32K configuration contains no latest preference instances, this category was aggregated only over the 128K and 1M configurations.
\begin{figure}[t]
    \centering
    \includegraphics[width=\columnwidth]{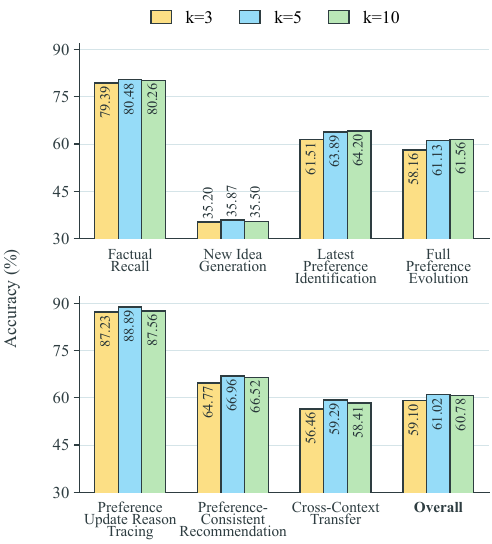}
    \caption{Ablation on key hyperparameters.}
    \label{fig:topk-sensitivity}
\end{figure}
As shown in Figure~\ref{fig:topk-sensitivity}, increasing $k$ from 3 to 5 raised overall accuracy from 59.10\% to 61.02\%, suggesting that a smaller retrieval depth may fail to cover the complementary evidence needed for user-state reconstruction. Increasing the depth further to $k=10$ slightly reduced overall accuracy to 60.78\%, indicating that further expanding the candidate set did not yield additional gains and may introduce redundant or weakly relevant memories. We therefore used $k=5$ as the main setting; it achieved the highest overall accuracy and the best results in five of the seven query categories.

\subsection{Efficiency Analysis}
\begin{table}[t]
\centering
\small
\setlength{\tabcolsep}{2pt}
\begin{tabular}{lccc}
\toprule
Method & Builds/100 $\downarrow$ & LLM Calls/100 $\downarrow$ & Tokens/Turn $\downarrow$ \\
\midrule
A-MEM & 100.0 & 201.2 & 3{,}317.9 \\
Mem0 & \textbf{25.6} & 380.7 & 6{,}401.9 \\
Zep & 100.0 & 331.8 & 13{,}337.3 \\
\textsc{QUMem} & 27.1 & \textbf{81.4} & \textbf{1{,}221.2} \\
\bottomrule
\end{tabular}
\caption{Normalized memory construction costs of different methods on PersonaMem.}
\label{tab:memory-construction-efficiency}
\end{table}

To compare memory-construction costs across methods when processing interaction histories of the same size, we define each user or assistant message as one dialogue turn and exclude system messages. PersonaMem contains 160{,}735 dialogue turns in total. Table~\ref{tab:memory-construction-efficiency} reports counts of memory constructions and generative LLM calls per 100 turns, together with the average number of construction tokens per turn, including both input and output tokens.

A-MEM and Zep trigger memory construction at every turn, resulting in high construction frequencies. Mem0 instead partitions each session into fixed chunks, subject to limits of 64 messages and 24,000 characters, thereby reducing the number of constructions. However, for each chunk, it first uses one LLM call to extract candidate memory units and then verifies each unit individually. Based on our experimental results, each construction requires nearly 15 LLM calls on average. Thus, a lower construction frequency does not translate into lower generative overhead, indicating that the number of constructions alone is insufficient to characterize memory-construction efficiency.

\textsc{QUMem} uses a local semantic-continuity classifier to form variable-length episodes and triggers memory construction only when an episode ends. For each episode, \textsc{QUMem} makes exactly three LLM calls to extract factual memories, preference memories, and transferable insights, respectively. This prevents the number of calls from increasing with the number of memory units. By combining a low construction frequency with a fixed number of LLM calls per construction,\textsc{QUMem} achieves the lowest LLM call frequency and token cost. These results indicate that \textsc{QUMem}'s efficiency gains arise from jointly reducing construction triggers and bounding the generative workload within each construction. The reported statistics cover only the cost of generative LLM calls during memory construction and exclude the inference cost of the local semantic-continuity classifier.

\section{Conclusion}
We introduce \textsc{QUMem}, a structured memory framework that combines Dynamic Episode Construction, Typed Memory Decomposition, and Query-Conditioned User-State Inference to infer task-relevant user states from long, evolving interaction histories. On PersonaMem, \textsc{QUMem} achieves the best performance among the evaluated methods, demonstrating the effectiveness of query-conditioned user-state inference for long-term personalization. These results highlight the value of preserving event context, distinguishing the roles of historical evidence, and performing task-driven user-state inference in long-term personalized memory systems.

\bibliography{aaai2027}

@misc{xu2025amemagenticmemoryllm,
      title={A-MEM: Agentic Memory for LLM Agents}, 
      author={Wujiang Xu and Zujie Liang and Kai Mei and Hang Gao and Juntao Tan and Yongfeng Zhang},
      year={2025},
      eprint={2502.12110},
      archivePrefix={arXiv},
      primaryClass={cs.CL},
      url={https://arxiv.org/abs/2502.12110}, 
}

@article{chhikara2025mem0,
  title={Mem0: Building production-ready ai agents with scalable long-term memory},
  author={Chhikara, Prateek and Khant, Dev and Aryan, Saket and Singh, Taranjeet and Yadav, Deshraj},
  journal={arXiv preprint arXiv:2504.19413},
  year={2025}
}

@misc{jiang2025knowmerespondme,
      title={Know Me, Respond to Me: Benchmarking LLMs for Dynamic User Profiling and Personalized Responses at Scale}, 
      author={Bowen Jiang and Zhuoqun Hao and Young-Min Cho and Bryan Li and Yuan Yuan and Sihao Chen and Lyle Ungar and Camillo J. Taylor and Dan Roth},
      year={2025},
      eprint={2504.14225},
      archivePrefix={arXiv},
      primaryClass={cs.CL},
      url={https://arxiv.org/abs/2504.14225}, 
}

@misc{chen2026knowubench,
      title={KnowU-Bench: Towards Interactive, Proactive, and Personalized Mobile Agent Evaluation},
      author={Tongbo Chen and Zhengxi Lu and Zhan Xu and Guocheng Shao and Shaohan Zhao and Fei Tang and Yong Du and Kaitao Song and Yizhou Liu and Yuchen Yan and Wenqi Zhang and Xu Tan and Weiming Lu and Jun Xiao and Yueting Zhuang and Yongliang Shen},
      year={2026},
      eprint={2604.08455},
      archivePrefix={arXiv},
      primaryClass={cs.AI},
      url={https://arxiv.org/abs/2604.08455},
}

@inproceedings{maharana-etal-2024-evaluating,
    title = "Evaluating Very Long-Term Conversational Memory of {LLM} Agents",
    author = "Maharana, Adyasha  and Lee, Dong-Ho  and Tulyakov, Sergey  and Bansal, Mohit  and Barbieri, Francesco  and Fang, Yuwei",
    booktitle = "Proceedings of the 62nd Annual Meeting of the Association for Computational Linguistics (Volume 1: Long Papers)",
    year = "2024",
    address = "Bangkok, Thailand",
    publisher = "Association for Computational Linguistics",
    url = "https://aclanthology.org/2024.acl-long.747/",
    doi = "10.18653/v1/2024.acl-long.747",
    pages = "13851--13870"
}

@inproceedings{tan-etal-2025-prospect,
    title = "In Prospect and Retrospect: Reflective Memory Management for Long-term Personalized Dialogue Agents",
    author = "Tan, Zhen  and
      Yan, Jun  and
      Hsu, I-Hung  and
      Han, Rujun  and
      Wang, Zifeng  and
      Le, Long  and
      Song, Yiwen  and
      Chen, Yanfei  and
      Palangi, Hamid  and
      Lee, George  and
      Iyer, Anand Rajan  and
      Chen, Tianlong  and
      Liu, Huan  and
      Lee, Chen-Yu  and
      Pfister, Tomas",
    editor = "Che, Wanxiang  and
      Nabende, Joyce  and
      Shutova, Ekaterina  and
      Pilehvar, Mohammad Taher",
    booktitle = "Proceedings of the 63rd Annual Meeting of the Association for Computational Linguistics (Volume 1: Long Papers)",
    month = jul,
    year = "2025",
    address = "Vienna, Austria",
    publisher = "Association for Computational Linguistics",
    url = "https://aclanthology.org/2025.acl-long.413/",
    doi = "10.18653/v1/2025.acl-long.413",
    pages = "8416--8439",
    ISBN = "979-8-89176-251-0"
}

@inproceedings{yue-etal-2026-hypermem,
    title = "{H}yper{M}em: Hypergraph Memory for Long-Term Conversations",
    author = "Yue, Juwei  and
      Hu, Chuanrui  and
      Sheng, Jiawei  and
      Zhou, Zuyi  and
      Zhang, Wenyuan  and
      Liu, Tingwen  and
      Guo, Li  and
      Deng, Yafeng",
    editor = "Liakata, Maria  and
      Moreira, Viviane P.  and
      Zhang, Jiajun  and
      Jurgens, David",
    booktitle = "Proceedings of the 64th Annual Meeting of the {A}ssociation for {C}omputational {L}inguistics (Volume 1: Long Papers)",
    month = jul,
    year = "2026",
    address = "San Diego, California, United States",
    publisher = "Association for Computational Linguistics",
    url = "https://aclanthology.org/2026.acl-long.1627/",
    doi = "10.18653/v1/2026.acl-long.1627",
    pages = "35237--35254",
    ISBN = "979-8-89176-390-6"
}

@inproceedings{10.1145/3774904.3792089,
author = {Zhong, Yijie and Gao, Yunfan and Wang, Haofen},
title = {HingeMem: Boundary Guided Long-Term Memory with Query Adaptive Retrieval for Scalable Dialogues},
year = {2026},
isbn = {9798400723070},
publisher = {Association for Computing Machinery},
address = {New York, NY, USA},
url = {https://doi.org/10.1145/3774904.3792089},
doi = {10.1145/3774904.3792089},
booktitle = {Proceedings of the ACM Web Conference 2026},
pages = {3576–3587},
numpages = {12},
location = {United Arab Emirates},
series = {WWW '26}
}

@misc{qin2026memoryretrievalchangingpreferences,
      title={Memory Retrieval for Changing Preferences}, 
      author={Yuehan Qin and Li Li and Linxin Song and Wei Yang and Jiate Li and Yuqing Yang and Yue Zhao},
      year={2026},
      eprint={2606.02976},
      archivePrefix={arXiv},
      primaryClass={cs.CL},
      url={https://arxiv.org/abs/2606.02976}, 
}

@misc{yang2026ramemcontextualreinstatementlongterm,
      title={RaMem: Contextual Reinstatement for Long-term Agentic Memory}, 
      author={Wei Yang and Bryce Kan and Shixuan Li and Li Li and Yuehan Qin and Jiate Li and Paul Bogdan and Jesse Thomason},
      year={2026},
      eprint={2606.22844},
      archivePrefix={arXiv},
      primaryClass={cs.AI},
      url={https://arxiv.org/abs/2606.22844}, 
}

@misc{chao2026stalellmagentsknow,
      title={STALE: Can LLM Agents Know When Their Memories Are No Longer Valid?}, 
      author={Hanxiang Chao and Yihan Bai and Rui Sheng and Tianle Li and Yushi Sun},
      year={2026},
      eprint={2605.06527},
      archivePrefix={arXiv},
      primaryClass={cs.CL},
      url={https://arxiv.org/abs/2605.06527}, 
}

@misc{yao2023reactsynergizingreasoningacting,
      title={ReAct: Synergizing Reasoning and Acting in Language Models}, 
      author={Shunyu Yao and Jeffrey Zhao and Dian Yu and Nan Du and Izhak Shafran and Karthik Narasimhan and Yuan Cao},
      year={2023},
      eprint={2210.03629},
      archivePrefix={arXiv},
      primaryClass={cs.CL},
      url={https://arxiv.org/abs/2210.03629}, 
}

@article{wang2023voyager,
  title={Voyager: An open-ended embodied agent with large language models},
  author={Wang, Guanzhi and Xie, Yuqi and Jiang, Yunfan and Mandlekar, Ajay and Xiao, Chaowei and Zhu, Yuke and Fan, Linxi and Anandkumar, Anima},
  journal={arXiv preprint arXiv:2305.16291},
  year={2023}
}

@misc{qin2025uitarspioneeringautomatedgui,
      title={UI-TARS: Pioneering Automated GUI Interaction with Native Agents}, 
      author={Yujia Qin and Yining Ye and Junjie Fang and Haoming Wang and Shihao Liang and Shizuo Tian and Junda Zhang and Jiahao Li and Yunxin Li and Shijue Huang and Wanjun Zhong and Kuanye Li and Jiale Yang and Yu Miao and Woyu Lin and Longxiang Liu and Xu Jiang and Qianli Ma and Jingyu Li and Xiaojun Xiao and Kai Cai and Chuang Li and Yaowei Zheng and Chaolin Jin and Chen Li and Xiao Zhou and Minchao Wang and Haoli Chen and Zhaojian Li and Haihua Yang and Haifeng Liu and Feng Lin and Tao Peng and Xin Liu and Guang Shi},
      year={2025},
      eprint={2501.12326},
      archivePrefix={arXiv},
      primaryClass={cs.AI},
      url={https://arxiv.org/abs/2501.12326}, 
}

@misc{rasmussen2025zeptemporalknowledgegraph,
      title={Zep: A Temporal Knowledge Graph Architecture for Agent Memory}, 
      author={Preston Rasmussen and Pavlo Paliychuk and Travis Beauvais and Jack Ryan and Daniel Chalef},
      year={2025},
      eprint={2501.13956},
      archivePrefix={arXiv},
      primaryClass={cs.CL},
      url={https://arxiv.org/abs/2501.13956}, 
}

@article{pan2025memory,
  title={On memory construction and retrieval for personalized conversational agents},
  author={Pan, Zhuoshi and Wu, Qianhui and Jiang, Huiqiang and Luo, Xufang and Cheng, Hao and Li, Dongsheng and Yang, Yuqing and Lin, Chin-Yew and Zhao, H Vicky and Qiu, Lili and others},
  journal={arXiv preprint arXiv:2502.05589},
  year={2025}
}

@inproceedings{park-etal-2026-learning,
    title = "Learning to Retrieve User History and Generate User Profiles for Personalized Persuasiveness Prediction",
    author = "Park, Sejun  and
      Park, Yoonah  and
      Lim, Jongwon  and
      Jo, Yohan",
    editor = "Liakata, Maria  and
      Moreira, Viviane P.  and
      Zhang, Jiajun  and
      Jurgens, David",
    booktitle = "Findings of the {A}ssociation for {C}omputational {L}inguistics: {ACL} 2026",
    month = jul,
    year = "2026",
    address = "San Diego, California, United States",
    publisher = "Association for Computational Linguistics",
    url = "https://aclanthology.org/2026.findings-acl.858/",
    doi = "10.18653/v1/2026.findings-acl.858",
    pages = "17338--17359",
    ISBN = "979-8-89176-395-1"
}

@inproceedings{van-etal-2026-memorai,
    title = "{M}em{ORAI}: Memory Organization and Retrieval via Adaptive Graph Intelligence for {LLM} Conversational Agents",
    author = "Van, Hung Pham  and
      Hieu, Nguyen Manh  and
      Tuan, Khang Pham Tran  and
      Hai, Nam Le  and
      Van, Linh Ngo  and
      Diep, Nguyen Thi Ngoc  and
      Le, Trung",
    editor = "Liakata, Maria  and
      Moreira, Viviane P.  and
      Zhang, Jiajun  and
      Jurgens, David",
    booktitle = "Findings of the {A}ssociation for {C}omputational {L}inguistics: {ACL} 2026",
    month = jul,
    year = "2026",
    address = "San Diego, California, United States",
    publisher = "Association for Computational Linguistics",
    url = "https://aclanthology.org/2026.findings-acl.1408/",
    doi = "10.18653/v1/2026.findings-acl.1408",
    pages = "28235--28253",
    ISBN = "979-8-89176-395-1"
}

@article{lewis2020retrieval,
  title={Retrieval-augmented generation for knowledge-intensive nlp tasks},
  author={Lewis, Patrick and Perez, Ethan and Piktus, Aleksandra and Petroni, Fabio and Karpukhin, Vladimir and Goyal, Naman and K{\"u}ttler, Heinrich and Lewis, Mike and Yih, Wen-tau and Rockt{\"a}schel, Tim and others},
  journal={Advances in neural information processing systems},
  volume={33},
  pages={9459--9474},
  year={2020}
}

@inproceedings{ma-etal-2023-query,
    title = "Query Rewriting in Retrieval-Augmented Large Language Models",
    author = "Ma, Xinbei  and
      Gong, Yeyun  and
      He, Pengcheng  and
      Zhao, Hai  and
      Duan, Nan",
    editor = "Bouamor, Houda  and
      Pino, Juan  and
      Bali, Kalika",
    booktitle = "Proceedings of the 2023 Conference on Empirical Methods in Natural Language Processing",
    month = dec,
    year = "2023",
    address = "Singapore",
    publisher = "Association for Computational Linguistics",
    url = "https://aclanthology.org/2023.emnlp-main.322/",
    doi = "10.18653/v1/2023.emnlp-main.322",
    pages = "5303--5315"
}

@inproceedings{lee-etal-2024-planrag,
    title = "{P}lan{RAG}: A Plan-then-Retrieval Augmented Generation for Generative Large Language Models as Decision Makers",
    author = "Lee, Myeonghwa  and
      An, Seonho  and
      Kim, Min-Soo",
    editor = "Duh, Kevin  and
      Gomez, Helena  and
      Bethard, Steven",
    booktitle = "Proceedings of the 2024 Conference of the North American Chapter of the Association for Computational Linguistics: Human Language Technologies (Volume 1: Long Papers)",
    month = jun,
    year = "2024",
    address = "Mexico City, Mexico",
    publisher = "Association for Computational Linguistics",
    url = "https://aclanthology.org/2024.naacl-long.364/",
    doi = "10.18653/v1/2024.naacl-long.364",
    pages = "6537--6555"
}

@inproceedings{petcu-etal-2026-query,
    title = "Query Decomposition for {RAG}: Balancing Exploration-Exploitation",
    author = "Petcu, Roxana  and
      Murray, Kenton  and
      Khashabi, Daniel  and
      Kanoulas, Evangelos  and
      Rijke, Maarten de  and
      Lawrie, Dawn  and
      Duh, Kevin",
    editor = "Demberg, Vera  and
      Inui, Kentaro  and
      Marquez, Llu{\'i}s",
    booktitle = "Proceedings of the 19th Conference of the {E}uropean Chapter of the {A}ssociation for {C}omputational {L}inguistics (Volume 1: Long Papers)",
    month = mar,
    year = "2026",
    address = "Rabat, Morocco",
    publisher = "Association for Computational Linguistics",
    url = "https://aclanthology.org/2026.eacl-long.322/",
    doi = "10.18653/v1/2026.eacl-long.322",
    pages = "6857--6871",
    ISBN = "979-8-89176-380-7"
}

@inproceedings{wang-etal-2023-large,
    title = "Large Language Models as Source Planner for Personalized Knowledge-grounded Dialogues",
    author = "Wang, Hongru  and
      Hu, Minda  and
      Deng, Yang  and
      Wang, Rui  and
      Mi, Fei  and
      Wang, Weichao  and
      Wang, Yasheng  and
      Kwan, Wai-Chung  and
      King, Irwin  and
      Wong, Kam-Fai",
    editor = "Bouamor, Houda  and
      Pino, Juan  and
      Bali, Kalika",
    booktitle = "Findings of the Association for Computational Linguistics: EMNLP 2023",
    month = dec,
    year = "2023",
    address = "Singapore",
    publisher = "Association for Computational Linguistics",
    url = "https://aclanthology.org/2023.findings-emnlp.641/",
    doi = "10.18653/v1/2023.findings-emnlp.641",
    pages = "9556--9569"
}

@inproceedings{chen-etal-2025-towards-omni,
    title = "Towards Omni-{RAG}: Comprehensive Retrieval-Augmented Generation for Large Language Models in Medical Applications",
    author = "Chen, Zhe  and
      Liao, Yusheng  and
      Jiang, Shuyang  and
      Wang, Pingjie  and
      Guo, YiQiu  and
      Wang, Yanfeng  and
      Wang, Yu",
    editor = "Che, Wanxiang  and
      Nabende, Joyce  and
      Shutova, Ekaterina  and
      Pilehvar, Mohammad Taher",
    booktitle = "Proceedings of the 63rd Annual Meeting of the Association for Computational Linguistics (Volume 1: Long Papers)",
    month = jul,
    year = "2025",
    address = "Vienna, Austria",
    publisher = "Association for Computational Linguistics",
    url = "https://aclanthology.org/2025.acl-long.742/",
    doi = "10.18653/v1/2025.acl-long.742",
    pages = "15285--15309",
    ISBN = "979-8-89176-251-0"
}

@inproceedings{guo-etal-2026-deepsieve,
    title = "{D}eep{S}ieve: Information Sieving via {LLM}-as-a-Knowledge-Router",
    author = "Guo, Minghao  and
      Zeng, Qingcheng  and
      Zhao, Xujiang  and
      Liu, Yanchi  and
      Yu, Wenchao  and
      Du, Mengnan  and
      Chen, Haifeng  and
      Cheng, Wei",
    editor = "Demberg, Vera  and
      Inui, Kentaro  and
      Marquez, Llu{\'i}s",
    booktitle = "Findings of the {A}ssociation for {C}omputational {L}inguistics: {EACL} 2026",
    month = mar,
    year = "2026",
    address = "Rabat, Morocco",
    publisher = "Association for Computational Linguistics",
    url = "https://aclanthology.org/2026.findings-eacl.160/",
    doi = "10.18653/v1/2026.findings-eacl.160",
    pages = "3054--3077",
    ISBN = "979-8-89176-386-9"
}

@inproceedings{yu-etal-2024-chain,
    title = "Chain-of-Note: Enhancing Robustness in Retrieval-Augmented Language Models",
    author = "Yu, Wenhao  and
      Zhang, Hongming  and
      Pan, Xiaoman  and
      Cao, Peixin  and
      Ma, Kaixin  and
      Li, Jian  and
      Wang, Hongwei  and
      Yu, Dong",
    editor = "Al-Onaizan, Yaser  and
      Bansal, Mohit  and
      Chen, Yun-Nung",
    booktitle = "Proceedings of the 2024 Conference on Empirical Methods in Natural Language Processing",
    month = nov,
    year = "2024",
    address = "Miami, Florida, USA",
    publisher = "Association for Computational Linguistics",
    url = "https://aclanthology.org/2024.emnlp-main.813/",
    doi = "10.18653/v1/2024.emnlp-main.813",
    pages = "14672--14685"
}

@inproceedings{xiao-etal-2026-mass,
    title = "{MASS}-{RAG}: Multi-Agent Synthesis Retrieval-Augmented Generation",
    author = "Xiao, Xingchen  and
      Huang, Heyan  and
      Liu, Runheng  and
      Xie, Jincheng",
    editor = "Liakata, Maria  and
      Moreira, Viviane P.  and
      Zhang, Jiajun  and
      Jurgens, David",
    booktitle = "Findings of the {A}ssociation for {C}omputational {L}inguistics: {ACL} 2026",
    month = jul,
    year = "2026",
    address = "San Diego, California, United States",
    publisher = "Association for Computational Linguistics",
    url = "https://aclanthology.org/2026.findings-acl.480/",
    doi = "10.18653/v1/2026.findings-acl.480",
    pages = "9865--9883",
    ISBN = "979-8-89176-395-1"
}

@article{zhang2025survey,
  title={A survey on the memory mechanism of large language model-based agents},
  author={Zhang, Zeyu and Dai, Quanyu and Bo, Xiaohe and Ma, Chen and Li, Rui and Chen, Xu and Zhu, Jieming and Dong, Zhenhua and Wen, Ji-Rong},
  journal={ACM Transactions on Information Systems},
  volume={43},
  number={6},
  pages={1--47},
  year={2025},
  publisher={ACM New York, NY}
}

% Include the reproducibility checklist here if required by the venue.
% \input{ReproducibilityChecklist.tex}

\end{document}